# A Privacy-Preserving Domain Adversarial Federated learning for multi-site brain functional connectivity analysis

Yipu Zhang[1, 2 *], Likai Wang[3], Kuan-Jui Su[2], Aiying Zhang[4], Hao Zhu[2], Xiaowen Liu[2], Hui Shen[2], Vince D. Calhoun[5], Yuping Wang[6], Hongwen Deng[2 *].

[1] School of Energy and Electrical Engineering, Chang'an University, Xi'an 710064, China.

[2] Tulane Center for Biomedical Informatics and Genomics, School of Medicine, Tulane University, New Orleans, LA 70112, USA.

[3] School of Electronics and Control Engineering, Chang'an University, Xi'an 710064, China.

[4] School of Data Science, University of Virginia, Charlottesville, VA 22903, USA

[5] Tri-Institutional Center for Translational Research in Neuroimaging and Data Science (TReNDS) Georgia State University, Georgia Institute of Technology, Emory University, Atlanta, GA 30030, USA.

[6] Department of Biomedical Engineering, Tulane University, New Orleans, LA 70112, USA.

[*] Corresponding authors: yzhang76@tulane.edu and hdeng2@tulane.edu.

## Abstract

Resting-state functional magnetic resonance imaging (rs-fMRI) and its derived functional connectivity networks (FCNs) have become critical for understanding neurological disorders. However, collaborative analyses and the generalizability of models still face significant challenges due to privacy regulations and the non-IID (non-independent and identically distributed) property of multiple data sources. To mitigate these difficulties, we propose Domain Adversarial Federated Learning (DAFed), a novel federated deep learning framework specifically designed for non-IID fMRI data analysis in multi-site settings. DAFed addresses these challenges through feature disentanglement, decomposing the latent feature space into domain-invariant and domain-specific components, to ensure robust global learning while preserving local data specificity. Furthermore, adversarial training facilitates effective knowledge transfer between labeled and unlabeled datasets, while a contrastive learning module enhances the global representation of domain-invariant features. We evaluated DAFed on the diagnosis of autism spectrum disorder (ASD) and further validated its generalizability in the classification of Alzheimer's disease (AD), demonstrating its superior classification accuracy compared to state-of-the-art methods. Additionally, an enhanced Score-CAM module identifies key brain regions and functional connectivity significantly associated with ASD and mild cognitive impairment (MCI), respectively, uncovering shared neurobiological patterns across sites. These findings highlight the potential of DAFed to advance multi-site collaborative research in neuroimaging while protecting data confidentiality.

**Introduction**

Resting-state functional magnetic resonance imaging (rs-fMRI) has emerged as a powerful and non-invasive technique for detecting abnormal brain activity [1]. Functional connectivity networks (FCNs), derived from rs-fMRI data, quantify temporal correlations between functional interactions in different brain regions, which are extensively utilized in studies of neurological disorders and mental illnesses [2, 3].

Recently, deep learning approaches have shown remarkable potential in analyzing fMRI data and FCNs, enabling significant breakthroughs in understanding brain function [4, 5]. Despite significant advancements in deep learning models, concerns over patient privacy and legal restrictions limit data sharing across institutions. This limitation poses challenges to the reproducibility and generalizability of data-driven approaches across diverse datasets [6, 7]. A primary factor contributing to these challenges is data heterogeneity, which arises from various sources, such as differences in MRI scanner hardware, imaging acquisition protocols, and regional disparities in data samples [8, 9], leading to non-IID (non-independent and identically distributed) data. Although methods such as ComBat can correct batch effects [10, 77], Although methods such as ComBat can correct batch effects [10, 77], ComBat requires consistent feature dimensions across different sites and cannot address the data heterogeneity arising from observational discrepancies across multiple data sources.

On the other hand, federated learning (FL) is a decentralized machine learning approach that facilitates collaborative model training while ensuring data remains localized at its originating site. Widely adopted in multi-site neuroimaging research [11, 78, 79], FL enables sites to train models independently on their local data and contribute to a shared global model by exchanging only model parameters, preserving data privacy and security.

Some recent studies have attempted to address the non-IID issue in the FL framework. For example, FedProx introduced a proximal term into the optimization objective at each site, constraining the distance between local and global models [12]. Similarly, FedMA addressed the problem of randomized parameter arrangements in multi-center non-IID data using a Bayesian non-parametric approach for hierarchical matching and fusion of gradient information from layered network models [13]. Yao *et al*. proposed an unbiased gradient aggregation algorithm that utilized keep-trace gradient descent along with a gradient evaluation strategy [14]. In addition, FedBN introduced an effective method that employed local batch normalization to alleviate feature shifts before averaging models [15]. However, most of these methods rely on batch normalization locally, which often disrupts the underlying spatial and temporal relationships within the data and is not well-suited for high-dimensional fMRI feature learning [15, 16]. Moreover, these techniques are primarily designed for labeled datasets and fails to address practical challenges, such as the imbalanced sample sizes and unlabeled data commonly encountered in clinical datasets [17].

To address these issues, we draw inspirations from domain adaptation (DA) and use the concept of domain adversarial training within the federated learning framework [18], developing a novel Domain Adversarial Federated learning framework (DAFed) to address the differences in feature distribution across multiple fMRI connectivity datasets. Specifically, we design a feature extractor to learn the spatial-temporal characteristics of fMRI data within a latent space. Then we utilize feature disentanglement technique to decompose the features into domain-invariant and domain-specific components [19]. The domain-invariant component captures the common features across all sites. They are collaboratively updated through federated learning, and the domain-specific component retains each dataset's unique, localized information. By employing a multi-head attention mechanism to integrate these two components, the classification accuracy for each dataset can be effectively enhanced. Moreover, by incorporating objective loss into the parameter transmission process during federated learning, adversarial training facilitates the transfer of information learned from labeled to unlabeled data. In addition, we introduce contrastive learning to strengthen the similarity of both global domain-invariant features across multiple sites and the domain-invariant features unique to each site. To evaluate the effectiveness of our method, we tested DAFed on the multi-site cohort Autism Brain Imaging Data Exchange (ABIDE) [20] and further validated its generalizability using the multi-device dataset from the Alzheimer's Disease Neuroimaging Initiative (ADNI). Our results demonstrate that DAFed outperforms many deep learning methods and recently proposed federated learning approaches resulting in higher classification accuracy. Furthermore, using an improved Score - class activation map (Score-CAM) module, we identify common key brain regions and functional connections across multiple sites that are significantly associated with ASD and AD. This collaborative analysis highlights shared neurobiological patterns across datasets, enhancing our understanding of these brain disorders.

The primary contributions of this paper can be summarized as follows:

- We propose the DAFed, an end-to-end federated deep learning framework designed for analyzing non-IID fMRI data collected from diverse scanners or institutions without the need for data sharing.
- We develop a feature extractor specifically designed to capture the spatial-temporal characteristics of fMRI data, effectively utilizing the dynamic FCN and its network structure information.
- We design a deep learning network that integrates feature disentanglement with domain adversarial training and embedded it within the FL framework, facilitating collaborative learning across multiple labeled and unlabeled datasets.
- We incorporate contrastive learning into the learning process of domain-invariant component across multiple sites, further enhancing the similarity between the domain-invariant component of the global FL model and the local models of individual sites.

- We introduce an optimized Score-CAM mechanism to mitigate the impact of noisy gradients from privacy-preserving noise in federated learning, improving model interpretability and advancing biomarker discovery.

## Methodology

### Problem definition

Consider collaboratively training a global model from $K$ sites (clients) using their respective datasets $\{X_1, \ldots, X_K\}$, where each dataset $X_k$ ($k = 1, \ldots, K$) shares the same feature space and label while differs in samples and data preprocessing protocols. As a result, the feature distribution of each dataset is non-IID. In this setup, the labeled datasets are referred to as the source domains, and the unlabeled datasets are the target domains. We aim to train an efficient model in the federated learning setting without sharing data.

For clarity, let the source domain site be $X_{source}$ and target domain site be $X_{target}$ as follows:

$$X_{source} = \{(x_i, y_i)\}_{i=1}^{n} \sim \mathbb{P}_{source}$$

$$X_{target} = \{(x_i)\}_{i=1}^{n'} \sim \mathbb{P}_{target}$$

where $n$ and $n'$ represent the number of samples in $X_{source}$ and $X_{target}$, with $n + n' = N$. The variables $x_i$ and $y_i$ denote the features and labels of the $i$-th sample, respectively, where $\mathbb{P}_{source}$ and $\mathbb{P}_{target}$ are the probability distributions followed by $X_{source}$ and $X_{target}$, respectively. In the later section, we will extend such a two-site model to a multi-site federated learning framework.

### A. Spatial-Temporal Feature Generator

Functional connectivity, measured by Pearson's correlation, is derived from fMRI time series and forms a graph representation. This graph has been widely utilized to study connectivity patterns in the human brain. To capture rich spatial-temporal information from FC data, graph convolutional networks (GCNs) are employed for feature extraction at the outset of the proposed framework, as shown in **Fig. 1**.

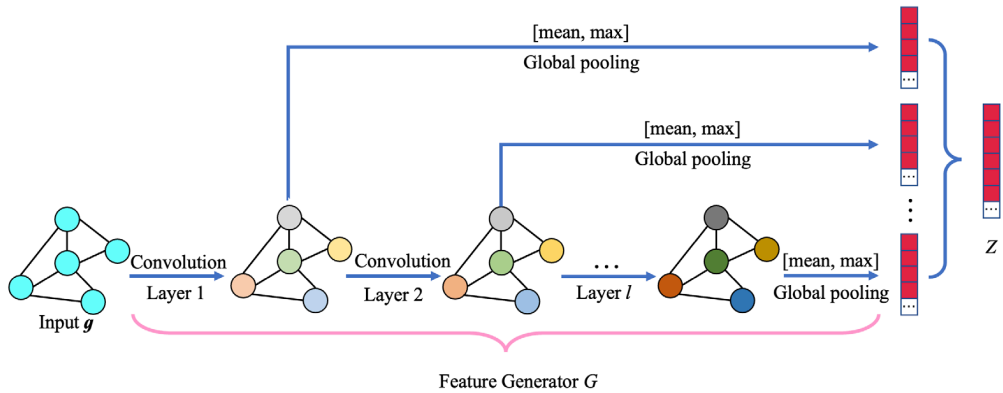

**Fig. 1**. Process of Feature Generation Based on Jumping Knowledge for GCN.

To capture temporal characteristics, we utilize dynamic functional connectivity (dFC) [80], modeled as an input graph $g = [v, e]$, where the node set $v$ represents regions of interest (ROIs) in the brain, and the edge set $e$ encodes the functional activation correlations between these ROIs. The graph convolutional layer is first used to aggregate neighboring node features, defined as follows:

$$H^{(l+1)} = \sigma\left(\widetilde{D}^{-\frac{1}{2}}(\widetilde{A})\widetilde{D}^{-\frac{1}{2}}H^{(l)}W^{(l)}\right) \quad (1)$$

where $\widetilde{A} = A + I$, $A$ is the adjacency matrix, $I$ is the identity matrix, $\widetilde{D}$ is the degree matrix, $W^{(l)}$ is the weight matrix for the $l$-th layer, $\sigma$ is the activation function, and $H^{(l)}$ represents the feature matrix at the $l$-th layer.

Next, a global pooling layer is employed to simplify the computation by applying global average pooling or global maximum pooling. We also integrate a Jumping Knowledge network [22] to concatenate pooled features from different convolution layers, enhancing the extraction of rich spatial information. Specifically, the global pooling layer reduces the feature matrix $H^{(l)}$ into a vector $z^{(l)}$. This "flattening" operation enables the extraction of a fixed-size representation. The output of the $l$-th global pooling layer is defined as follows:

$$z^{(l)} = \text{mean}(H^{(l)}) \parallel \text{max}(H^{(l)}) \quad (2)$$

where mean(·) and max(·) represent the average pooling and maximum pooling functions, respectively, used to aggregate the feature vectors of each node in the graph, and $\parallel$ denotes the concatenation. Finally, the embedding feature $Z$ is obtained by concatenating the outputs of all global pooling layers,

$$Z = z^{(1)} \parallel z^{(2)} \parallel \cdots z^{(L)} \quad (3)$$

where $L$ is the number of convolutional layers in the generator module.

### B. Representation Disentanglement

To address the feature distribution shift in multi-site datasets, a key hypothesis is that each dataset consists of both a domain-invariant component ($f_{di}$) and a domain-specific component ($f_{ds}$). Here, we utilize a feature disentangler $D$ to decompose the embedding feature $Z$ into two components: $f_{di} = D_{di}(Z)$ and $f_{ds} = D_{ds}(Z)$. We minimize the mutual information between $f_{di}$ and $f_{ds}$ to enhance this disentanglement, which is defined as follows:

$$I(f_{di}, f_{ds}) = \iint p(f_{di}, f_{ds}) \log \frac{p(f_{di}, f_{ds})}{p(f_{di})p(f_{ds})} df_{di} df_{ds} \quad (4)$$

where $p(f_{di}, f_{ds})$ is the joint probability distribution of $f_{di}$ and $f_{ds}$, and $p(f_{di})$, $p(f_{ds})$ are their respective marginal distributions. Although mutual information is a measure for capturing dependencies across distributions, it is only tractable for discrete variables. To estimate the mutual information for continuous variables, we use the Mutual Information Neural Estimator (MINE), which employs a neural network and Monte Carlo integration [23]. The mutual information loss is defined as follows:

$$\mathcal{L}_{MI}(f_{di}, f_{ds}) = \left| \frac{1}{n} \sum_{i=1}^{n} T_\varphi(p, q) - \log\left(\frac{1}{n} \sum_{i=1}^{n} e^{T_\varphi(p', q')}\right) \right| \tag{5}$$

where $(p, q)$ are sampled from the joint distribution $p(f_{di}, f_{ds})$, and $(p', q')$ are sampled independently from the product of marginals $p(f_{di})p(f_{ds})$. The neural network $T_\varphi$, parameterized by $\varphi$, approximates the mutual information between $f_{di}$ and $f_{ds}$.

### C. Domain Adversarial Training

To address the non-IID issue between the source and target domains, we utilize a domain adversarial neural network. It aligns the source domain with the heterogeneous target domain via adversarial training on the domain-invariant component and preserves the unique information within the local domain-specific component. Domain adversarial training is achieved by two discriminators. The classifier ($C$) is trained on the source domain to evaluate model performance, and the domain identifier ($DI$) is used to determine the origin of the features. The loss of $DI$ is defined as follows:

$$\mathcal{L}_{DI} = \frac{1}{N} \sum_{i=1}^{N} \left( d_i \log \frac{1}{DI\left(\left(D_{di}(G(x_i))\right)\right)} + (1 - d_i) \log \frac{1}{1 - DI\left(\left(D_{di}(G(x_i))\right)\right)} \right) \tag{6}$$

where $d_i$ is a binary domain indicator for the $i$-th sample. Specifically, $d_i = 0$ indicates that $x_i$ come from the source distribution $x_i \sim \mathbb{P}_{source}$, and $d_i = 1$ indicates it comes from the target distribution $x_i \sim \mathbb{P}_{target}$.

To enhance the performance of the classifier $C$, we integrate the domain-invariant and domain-specific components to form a unified feature. This approach retains the shared features across the two datasets while preserving their unique characteristics by employing an improved multi-head self-attention mechanism [24]:

$$f_F = \text{Softmax}\left(\frac{QK^T}{\sqrt{\phi_K}}\right) V \tag{7}$$

where $Q$, $K$ and $V$ represent the Query, Key, and Value matrices, respectively. The term $QK^T$ forms the similarity matrix of the domain-invariant and domain-specific components. This relationship is normalized by the Softmax function to produce a soft attention matrix, scaled by $(\phi_k)^{1/2}$ to regulate the magnitude. The weighted output $V$ is then used to generate the integrated features $f_F$. Note that we omit the Mask operation from the Scaled Dot-Product Attention mechanism, as it is primarily used in natural language processing tasks, which do not apply to our work.

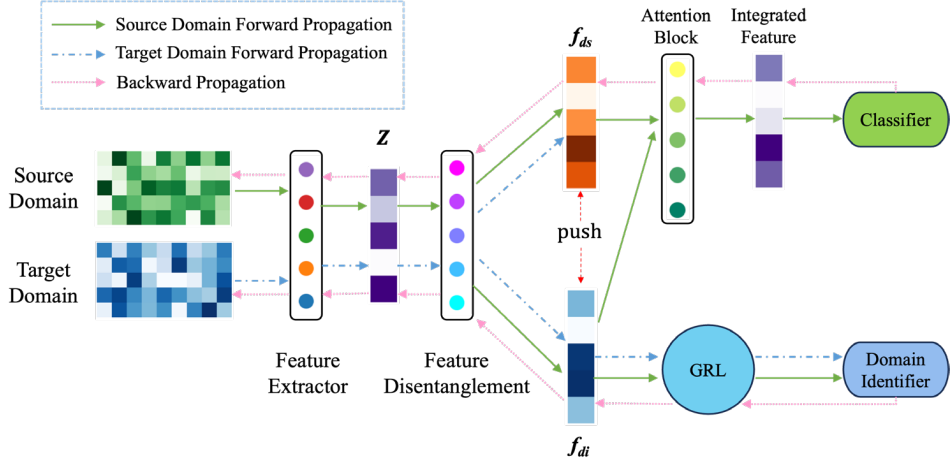

**Fig. 2.** Workflow of domain adversarial training.

For the sample $\{x_i, y_i\}$ in the source domain, the classification loss is given by

$$\mathcal{L}_C = \frac{1}{n}\sum_{i=1}^{n} y_i \log\frac{1}{C_{source}(f_F^i)} + (1 - y_i)\log\frac{1}{1 - C_{source}(f_F^i)} \tag{8}$$

where and $f_F^i$ is the integrated feature obtained from the multi-head self-attention mechanism. The classifier $C$ consists of MLP parameterized by $\theta_C$ and outputs the predicted label. Note that in this adversarial training, since the target domain is unlabeled, there is no need to calculate the classification loss. However, if the target domain is labeled, the classification loss should be included in the objective loss for the target domain.

As depicted in **Fig. 2**, we implement domain adversarial training to achieve federated learning by the following steps:

1. Initialize parameters $\Theta_{source} = \{\theta_G, \theta_D, \theta_{attention}, \theta_{DI}, \theta_C\}$ and perform forward propagation to calculate $L_{total\_source}$ in the source domain.
2. Send the parameters $\Theta_{source}$ and the loss $L_{total\_source}$ to the target domain.
3. In the target domain, input all target domain samples and perform forward propagation to calculate the loss $L_{total\_target}$ using the parameters $\Theta_{source}$.

4. The total objective loss $L = L_{total\_source} + L_{total\_target}$ is then used for backpropagation to update all parameters specified as $\Theta_{target}$.
   5. Send $\Theta_{target}$ back to the source domain, updating $\Theta_{source} = \Theta_{target}$ for subsequent iterations.

### D. Multi-site Federated Learning

Building on the Domain Adversarial Training described in Section II-C, we extend this model to a multi-site federated learning, as depicted in **Fig. 3**.

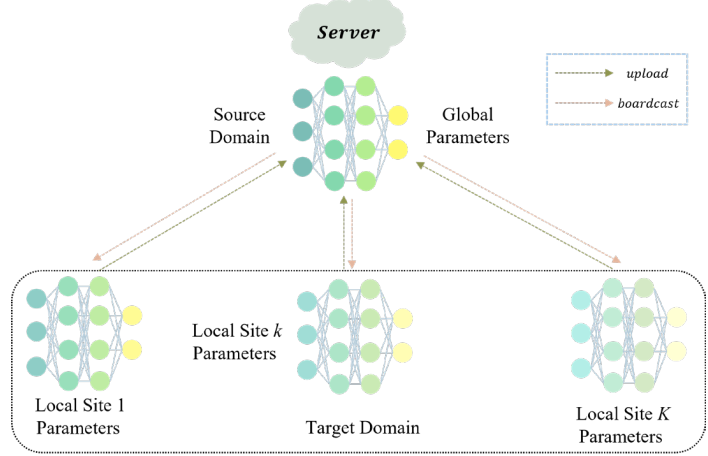

**Fig. 3**. Workflow of domain adversarial federated learning

To improve computational efficiency, we designate the labeled source domain as the central site, and the remaining unlabeled datasets are treated as local sites. The proposed algorithm operates as follows:

   1. At the central site (source domain), initialize with parameters $\Theta_{global}$ and perform forward propagation to compute the loss $L_{total\_source}$.
   2. The parameters $\Theta_{global}$ and the loss $L_{total\_source}$ are then broadcast to all local sites.
   3. Update each local parameter $\Theta_k = \Theta_{global}$ ($k = 1, \ldots, K$), and then, input the samples from each site $k$ to calculate the loss $L_{total\_target\_k}$ for each local site using the parameter $\Theta_k$.
   4. At each site $k$, use the total objective loss $L = L_{total\_target\_k} + L_{total\_source}$ to perform backpropagation and update $\Theta_k$.
   5. Add random noise to each updated $\Theta_k$ to protect privacy and then upload it back to the central site [81].
   6. At the central site, after receiving the $\Theta_k$ from each local site, optimize and update $\Theta_{global}$, completing one iteration.

### E. Contrastive Learning Module

Our proposed model aims to learn the domain-invariant component shared across both source and target domains and retain the domain-specific component within each domain. Recent advances in contrastive learning provide a new approach to enhance feature similarity by minimizing the distance between positive samples and maximizing the distance between negative samples. Inspired by this, we incorporate contrastive learning by aggregating the current local model with both the previous local and global models, further refining the domain-invariant component across all sites.

Specifically, at each site, we define the domain-invariant feature learned using the local parameters at the *t*-th iteration and the domain-invariant feature learned using the global parameters at the (*t* – 1)-th iteration as a positive sample pair. Conversely, the domain-invariant feature learned using the local parameters at the *t*-th iteration and the domain-invariant features learned using the local parameters during the previous (*t* – v) iterations are defined as negative sample pairs. **Fig. 4** illustrates the process of generating both positive and negative samples in the contrastive learning module.

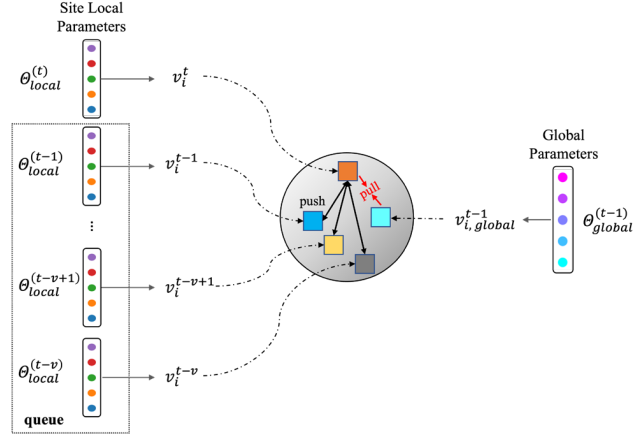

**Fig. 4**. Generation of positive and negative samples in the contrastive learning module

Overall, our goal is to make the global domain-invariant component obtained via federated learning to be more similar to the domain-invariant component learned at each local site. The contrastive loss is defined as follows:

$$L_{CL} = -\frac{1}{n}\sum_{i=1}^{n} \log \frac{e^{\frac{\text{sim}(\mathcal{V}_i^t, \mathcal{V}_{i,global}^{t-1})}{\tau}}}{e^{\frac{\text{sim}(\mathcal{V}_i^t, \mathcal{V}_{i,global}^{t-1})}{\tau}} + \sum_{j=1}^{v} e^{\frac{\text{sim}(\mathcal{V}_i^t, \mathcal{V}_i^{t-j})}{\tau}}} \quad (9)$$

where $\text{sim}(\cdot)$ represents the cosine similarity function, $\tau$ denotes the temperature parameter, and $t$ is the current iteration round. The domain-invariant component $\mathcal{V}_{i,global}^{t-1}$ is the *i*-th sample learned by the global parameters in the (*t* – 1) iteration,

$$\mathcal{V}_{i,global}^{t-1} = \left[D_{di_{global}}\left(G_{global}(x_i)\right)\right]^{(t-1)} \qquad (10)$$

Similarly, $\mathcal{V}_i^t$ represents the domain-invariant component of the *i*-th sample learned by the local parameters in the *t*-th iteration.

Finally, the total loss function of our model is defined as

$$L_{total} = L_C + \lambda_1 L_{MI} + \lambda_2 L_{CL} + \lambda_p L_{DI} \qquad (11)$$

where $L_C$ represents the classification loss, $L_{DI}$ is the domain classification loss, $L_{MI}$ denotes the mutual information loss, and $L_{CL}$ is the contrastive loss. The parameters $\lambda_1$ and $\lambda_2$ are regularization coefficients that control the weights of the mutual information loss and scale the contrastive learning weights, respectively. $\lambda_p$ is to balance the classification loss and the domain discrimination loss.

### F. Model Interpretation

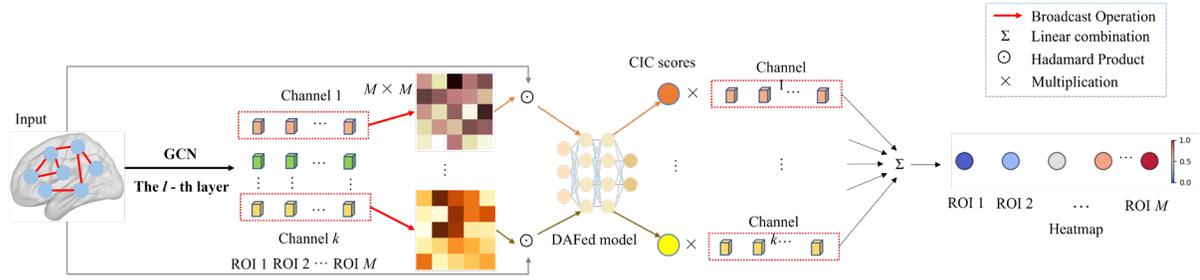

**Fig. 5.** The framework of the improved Score-CAM module. The process begins by extracting the activation map from the output of the *l*-th GCN layer. Each channel of the activation map corresponds to the same feature across different ROIs and is then broadcasted channel-by-channel. These broadcasted activation maps are used as masks to the input. For each masked input, the forward-pass score corresponding to the target class is computed. Finally, interpretable results are generated by linearly combining the activation maps with their associated score-based weights.

In Section II-D, to ensure privacy protection, we add random noise during the parameter transmission process, which results in noisy gradients. This noise interference directly impacts gradient-based interpretation methods in deep learning models [25, 26], such as Grad-CAM. To address this issue, we apply a Score-CAM method to models without a global pooling layer and operates independently of gradient information. We design an improved Score-CAM approach to reduce the influence of noise, enhancing the interpretability of the model, with which the most discriminative regions of interest (ROIs) can be identified. As illustrated in **Fig. 5**, we improve Score-CAM by replacing CNN-based convolutions with GCNs and incorporating the proposed model to compute the CIC score. Finally, the activation maps are generated, highlighting significant features relevant to the classification task.

The pseudocode of DAFed is summarized in **Algorithm 1**.

**Algorithm 1: The Domain Adversarial Federated Learning (DAFed)**

**Input**: Source domain dataset $X_{source}$ and label $y$, target domain datasets $X_k$ ($k = 1, 2, ..., K$), hyper-parameters $\lambda_1, \lambda_2, \lambda_p$ and learning rate $\mu$.
**Output**: Global model $\Theta_{global}$.

1. **Initialize** global model $\Theta_{global}$
2. **for** $t = 1$ to *Max-Iteration* **do**
3.     Calculate the objective loss $L_{total\_source} = L_C + \lambda_1 L_{MI} + \lambda_2 L_{CL} + \lambda_p L_{DI}$ by $\Theta_{global}^t$.
4.     **for** each local $k = 1$ to $K$ **do**
5.         $\Theta_k^t \leftarrow \Theta_{global}^t$
6.         calculate objective loss $L_{total\_target\_k} = \lambda_1 L_{MI} + \lambda_2 L_{CL} + \lambda_p L_{DI}$ by $\Theta_k^t$.
7.         update $\Theta_k^{t+1} \leftarrow \Theta_k^t - \mu \cdot \frac{\partial(L_{total\_source} + L_{total\_target\_k})}{\partial \Theta_k^t}$ and upload $\Theta_k^{t+1}$ to the server.
8.     **end for**
9.     $\Theta_{global}^{t+1} \leftarrow \text{avg}(\Theta_1^t, \Theta_2^t, \cdots, \Theta_K^t)$
10.     broadcast $\Theta_{global}^{t+1}$ to all $X_k$ ($k = 1, 2, ..., K$).
11.     **until** converge
12. **end for**
13. **return** Global model $\Theta_{global}$

## Experiments and results

### Data acquisition

We validate our method on two public datasets, *e.g.*, ABIDE and ADNI-3 [20, 21]. The ABIDE dataset gathers brain neuroimaging data from multiple institutions to enhance our understanding of the neural mechanisms underlying autism. For this study, we focused on resting-state fMRI data from the four sites: UM_1, NYU, USM, and UCLA_1. Subjects were selected based on the completeness of the time series data for each ROI. In total, 370 screened subjects were included across these four sites: NYU contributed 73 ASD cases and 94 normal controls (NC), UM_1 has 43 ASD and 45 NC, USM contributed 33 ASD and 19 NC, and UCLA_1 contributed 37 ASD and 26 NC. To further assess the consistency of our method across different MRI scanners, we used resting-state fMRI data from the ADNI-3 study, collected with three distinct MRI scanners: 3.0T GE, 3.0T Siemens, and 3.0T Philips. These three scanners provided a total of 844 screened subjects, with the Philips scanner contributing 111 MCI and 62 NC, the Siemens scanner contributing 129 MCI and 159 NC, and the GE scanner contributing 186 MCI and 197 NC. Data from each site or scanner were treated as independent datasets with no data sharing. In addition, due to a lack of sufficient data, we applied sliding windows to truncate the raw fMRI time series, following the experimental setup in [6]. The data characteristics and MRI scanner parameters for both the ABIDE and ADNI-3 datasets are summarized in **Table 1** and **Table 2**, respectively.

Table 1. Data characteristics and MRI scanner parameters of the ABIDE.

| | | F/M | Age | IQ | ASD/NC | fMRI Frames | Overlapping Truncation |
|---|---|---|---|---|---|---|---|
| NYU | ASD | 8/65 | 14.7±7.1 | 107.4±16.5 | 73/94 | 176 | 157 |
| | NC | 25/69 | 15.2±5.9 | 112.6±13.5 | | | |
| | | Echo Time | Repetition Time | Slice Number | Scanner (Field Strength) | Slice Thickness | |
| | fMRI | 15 ms | 2000 ms | 33 | Siemens Allegra (3.0 tesla) | 4.0 mm | |
| | | F/M | Age | IQ | ASD/NC | fMRI Frames | Overlapping Truncation |
| UM_1 | ASD | 7/36 | 12.4±2.2 | 102.8±18.8 | 43/45 | 296 | 277 |
| | NC | 13/32 | 14.1±3.4 | 106.7±9.6 | | | |
| | | Echo Time | Repetition Time | Slice Number | Scanner (Field Strength) | Slice Thickness | |
| | fMRI | 30 ms | 2000 ms | 40 | GE Signa (3.0 tesla) | 3.0 mm | |
| | | F/M | Age | IQ | ASD/NC | fMRI Frames | Overlapping Truncation |
| USM | ASD | 0/33 | 22.9±7.3 | 99.8±16.4 | 33/19 | 236 | 217 |
| | NC | 0/19 | 20.8±8.2 | 117.1±14.4 | | | |
| | | Echo Time | Repetition Time | Slice Number | Scanner (Field Strength) | Slice Thickness | |
| | fMRI | 28 ms | 2000 ms | 40 | Siemens Trio Tim (3.0 tesla) | 3.0 mm | |
| | | F/M | Age | IQ | ASD/NC | fMRI Frames | Overlapping Trunc |
| UCLA_1 | ASD | 6/31 | 13.0±2.7 | 103.5±13.5 | 37/26 | 116 | 97 |
| | NC | 4/22 | 13.4±2.3 | 104.9±10.4 | | | |
| | | Echo Time | Repetition Time | Slice Number | Scanner (Field Strength) | Slice Thickness | |
| | fMRI | 28 ms | 3000 ms | 34 | Siemens Magnetom TrioTim (3.0 tesla) | 4.0 mm | |

Table 2. Data characteristics and MRI scanner parameters of the ADNI-3.

|  |  | F/M | Age | IQ | MCI/NC | fMRI Frames | Overlapping Truncation |
|---|---|---|---|---|---|---|---|
| Philips | MCI | 46/65 | 75.83±7.81 | / | 111/62 | 187 | 168 |
|  | NC | 34/28 | 75.10±7.23 | / |  |  |  |
|  |  | Echo Time | Repetition Time | Slice Number | Scanner (Field Strength) | Slice Thickness | |
|  | fMRI | 30 ms | 3000 ms | 48 | Philips (3.0 tesla) | 3.4 mm | |
| Siemens |  | F/M | Age | IQ | MCI/NC | fMRI Frames | Overlapping Truncation |
|  | MCI | 47/82 | 73.48±6.98 | / | 129/159 | 200 | 181 |
|  | NC | 100/59 | 73.77±8.77 | / |  |  |  |
|  |  | Echo Time | Repetition Time | Slice Number | Scanner (Field Strength) | Slice Thickness | |
|  | fMRI | 30 ms | 3000 ms | 24 | Siemens (3.0 tesla) | 3.4 mm | |
| GE |  | F/M | Age | IQ | MCI/NC | fMRI Frames | Overlapping Truncation |
|  | MCI | 75/112 | 75.91±7.48 | / | 186/197 | 190 | 171 |
|  | NC | 106/91 | 74.06±7.70 | / |  |  |  |
|  |  | Echo Time | Repetition Time | Slice Number | Scanner (Field Strength) | Slice Thickness | |
|  | fMRI | 30 ms | 3000 ms | 48 | GE (3.0 tesla) | 3.4mm | |

**Data preprocessing**

The task performed on the ABIDE dataset was to classify subjects as ASD and NC. The raw fMRI images were preprocessed using the CPAC pipeline [27]. Primary steps included band-pass filtering (0.01 - 0.1 Hz) without global signal regression, and parcellation into 111 regions of interest (ROIs) using the Harvard-Oxford (HO) atlas. After preprocessing, we computed Pearson's correlation matrix from the slicing time series of ROI using a sliding window (window size = 20) to capture dynamic functional connectivity. Then the Fisher transformation is applied to Pearson's correlation matrix, and the feature matrix has dimensions of $111 \times 111$ for each sample [28-30].

The task we performed on the ADNI datasets was to identify MCI and normal control NC. All the resting-state fMRI (rs-fMRI) data from ADNI-3 were preprocessed using DPARSF [31]. Briefly, the preprocessing steps were as follows: the first 10 volumes of the functional images were discarded to ensure magnetization equilibrium. slice timing and head motion correction were then performed followed by normalization to the Montreal Neurological Institute (MNI) template and resampling to an isotropic voxel size of 3 mm. Additionally, spatial smoothing was conducted using a 4-mm full-width at half maximum (FWHM) Gaussian kernel. Detrending and bandpass filtering (0.01–0.1 Hz) were then applied, after which we regressed out covariates, including the six head motion parameters, white matter (WM), cerebrospinal fluid (CSF), and global signal, to minimize the influence of these confounding signals. Finally, time series data were extracted from brain regions based on the Anatomical Automatic Labeling (AAL) atlas, which included 116 ROIs. The functional connectivity of each sample was calculated in the same way as in ABIDE.

**Parameters setup**

The architecture of our proposed DAFed method is shown in **Table 3**. Specifically, we configured the number of graph convolution layers to 4. The number of heads in the self-attention mechanism was set to 8, enabling the model to capture diverse information across layers. In the contrastive learning module, the queue length was fixed at 5 to increase the number of negative samples.

Table 3. Model architecture of the DAFed model.

| Layer | Configuration |
|---|---|
| (1) Spatial-Temporal Feature Generator | |
| 1 | GCN (number of ROIs,128), BN, ReLU |
| 2 | Dropout (0.1), GCN (128,64), BN, ReLU |
| 3 | Dropout (0.1), GCN (64,32), BN, ReLU |
| 4 | Dropout (0.1), GCN (32,16), BN, ReLU |
| (2) Feature Disentanglement | |
| 1 | MLP (480,256), BN, ReLU |
| 2 | Dropout (0.2), MLP (256,128), BN, ReLU |
| (3) Improved Multi-head Self-attention Mechanism | |
| 1 | Multi-head Self-attention (hid_dim=128, heads=8, dropout=0) |
| (4) Mutual Information Neural Estimator | |
| 1 | MLP (128,32), BN, LeakyReLU |
| 2 | MLP (32,1) |
| (5) Domain Identifier | |
| 1 | MLP (128,160), BN, ReLU |
| 2 | Dropout (0.5), MLP (160,2), Softmax |
| (6) Classifier | |
| 1 | MLP (256,320), BN, ReLU |
| 2 | Dropout (0.5), MLP (320,2), Softmax |

In addition, the hyperparameters $\lambda_1$ and $\lambda_2$ in the total loss function were optimized through grid search and finalized as 1 and 0.1, respectively. $\lambda_P$ was defined as

$$\lambda_p = \frac{2}{1 + exp(-\gamma \cdot p)} - 1 \qquad (12)$$

This represents a monotonically increasing convex function that grows with the number of iterations, ranging between [0,1] [32], where $\gamma$ is a configurable parameter set to 10, and $p$ denotes the ratio of the current number of iterations to the total number of iterations. To preserve privacy, we added Gaussian noise $\varepsilon_n \sim N(0, \alpha\sigma)$ to the local model, where $\sigma$ is the standard deviation of the local model weights, and $\alpha$ represents the noise level. In our implementation, we set $\alpha = 0.01$. Due to differences in the ROI templates of the two datasets, we employed two learning rates. For the ABIDE dataset, the learning rate was initialized at 0.0001 and followed a two-step process, beginning with a warm-up phase before applying decay. For the ADNI dataset, the initial learning rate was set to 0.01 with a direct decay approach [33,34]. The Adam optimizer was used for optimization, and the batch size was set to 1/16 of the training set size. All experiments were conducted using PyTorch and executed on an NVIDIA 4080 GPU.

## Results

In the experiments, we utilized both the ABIDE and ADNI datasets to evaluate the classification performance of our proposed DAFed in comparison to several competing methods. The classification task for the ABIDE dataset focused on distinguishing between ASD and NC, while for the ADNI dataset, the task involved distinguishing between MCI and NC, with NC considered the positive class in both cases. The competing methods were divided into two categories: non-federated learning approaches, such as MLP and GCN [35], and federated learning-based approach, including Fed-Avg [16], Fed-MoE [6], Fed-Align [6], and FedCL [9]. Additionally, to evaluate the performance of the proposed method in domain adaptation, we employed two strategies: (1) without target domain labels (DAFed_U) and (2) with target domain labels (DAFed_L). A 5-fold cross-validation (CV) approach was used to evaluate the classification accuracy of all models.

**Table 4**. Comparison of classification accuracy on the ABIDE dataset

| Site / Model | NYU | UCLA_1 | UM_1 | USM | Average |
|---|---|---|---|---|---|
| MLP | 0.609 ± 0.103 | 0.545 ± 0.133 | 0.593 ± 0.191 | 0.689 ± 0.192 | 0.609 |
| GCN | 0.684 ± 0.057 | 0.655 ± 0.052 | 0.671 ± 0.063 | 0.704 ± 0.077 | 0.679 |
| FedAvg | 0.667 ± 0.075 | 0.719 ± 0.121 | 0.671 ± 0.106 | 0.618 ± 0.176 | 0.669 |
| FedMoE | 0.672 ± 0.126 | 0.741 ± 0.151 | 0.716 ± 0.142 | 0.773 ± 0.173 | 0.726 |
| FedAlign | 0.689 ± 0.077 | 0.636 ± 0.127 | 0.704 ± 0.107 | 0.733 ± 0.134 | 0.69 |
| FedCL | *0.707 ± 0.087* | 0.733 ± 0.107 | 0.648 ± 0.186 | 0.749 ± 0.241 | 0.709 |
| DAFed_U | 0.678 ± 0.070 | *0.75 ± 0.061* | *0.738 ± 0.120* | **0.847 ± 0.086** | *0.753* |
| DAFed_L | **0.735 ± 0.036** | **0.779 ± 0.052** | **0.741 ± 0.047** | *0.799 ± 0.035* | **0.764** |

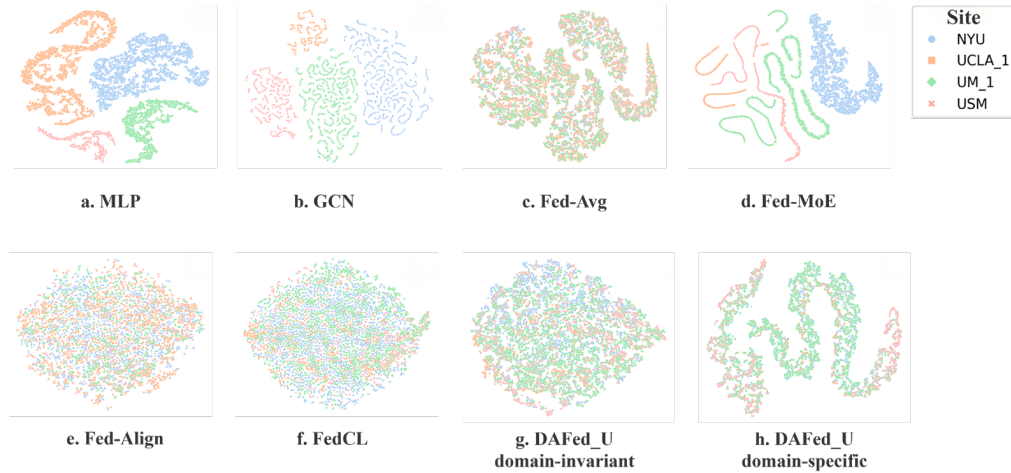

**Fig. 6.** Comparison of latent space visualization by different models on ABIDE.

**Table 4** summarizes the classification performance of comparative methods on the ABIDE dataset. The results show that the two non-federated learning methods achieved relatively low average accuracy, with 0.609 for MLP and 0.679 for GCN. To better understand whether these methods could capture common domain-invariant features, we utilized t-SNE [36, 82] to visualize the latent space embedded by the first fully connected layer of MLP and GCN, as shown in **Fig. 6a** and **6b**. The visualizations show that these non-federated learning models failed to learn robust common domain-invariant features, as features from each site are independently distributed.

For federated learning-based methods, the average accuracy scores were 0.669 for Fed-Avg, 0.672 for Fed-MoE, 0.689 for Fed-Align, and 0.707 for FedCL. The t-SNE visualizations in **Fig. 6c** to **6f** further highlight these differences. Fed-Avg partially captured common domain-invariant features, but the features still formed site-specific clusters. Fed-MoE did not effectively learn common domain-invariant features, it performed better than non-federated learning models, with features from different sites interwoven within the same latent space. Fed-Align and FedCL improved by successfully learning common domain-invariant features with site-specific features dispersed throughout the latent space. Since both non-federated learning methods have achieved better results on USM, and the feature distribution of USM appears relatively compact in Figures 6a and 6b, we designate USM as the source domain in DAFed. As shown in **Table 4**, our method achieves the best performance across all four datasets. The DAFed_L utilizing label information obtains the highest average accuracy of 0.764, and the unlabeled version DAFed_U achieves the second-highest average accuracy of 0.753. Moreover, compared to other methods, DAFed also achieves a smaller standard deviation, indicating more stable performance. **Fig. 6g** and **6h** further demonstrate that our method learns robust common domain-invariant features and enables domain-specific features to exhibit a certain level of aggregated distribution.

Similarly, we compared the performance of these methods on the ADNI-3 dataset, as shown in **Table 5**. Two non-federated learning methods achieved an average classification accuracy of 0.67 for MLP and 0.692 for GCN, respectively. After visualization, these two methods produced distinct feature distributions for different scanners, as shown in **Fig. 7a** and **7b**.

Table 5. Comparison of classification accuracy on the ADNI-3 dataset

| Site / Model | Philips | Siemens | GE | Average |
|---|---|---|---|---|
| MLP | 0.665 ± 0.102 | 0.596 ± 0.115 | 0.749 ± 0.099 | 0.67 |
| GCN | 0.722 ± 0.093 | 0.604 ± 0.037 | *0.749 ± 0.055* | 0.692 |
| FedAvg | *0.727 ± 0.048* | 0.586 ± 0.114 | 0.746 ± 0.039 | 0.686 |
| FedMoE | 0.653 ± 0.122 | 0.573 ± 0.031 | 0.721 ± 0.061 | 0.649 |
| FedAlign | 0.664 ± 0.081 | 0.577 ± 0.069 | 0.715 ± 0.067 | 0.652 |
| FedCL | 0.7 ± 0.106 | 0.631 ± 0.078 | 0.697 ± 0.097 | 0.676 |
| DAFed_U | 0.715 ± 0.039 | *0.635 ± 0.072* | 0.731 ± 0.031 | *0.694* |
| DAFed_L | **0.729 ± 0.101** | **0.671 ± 0.075** | **0.754 ± 0.047** | **0.718** |

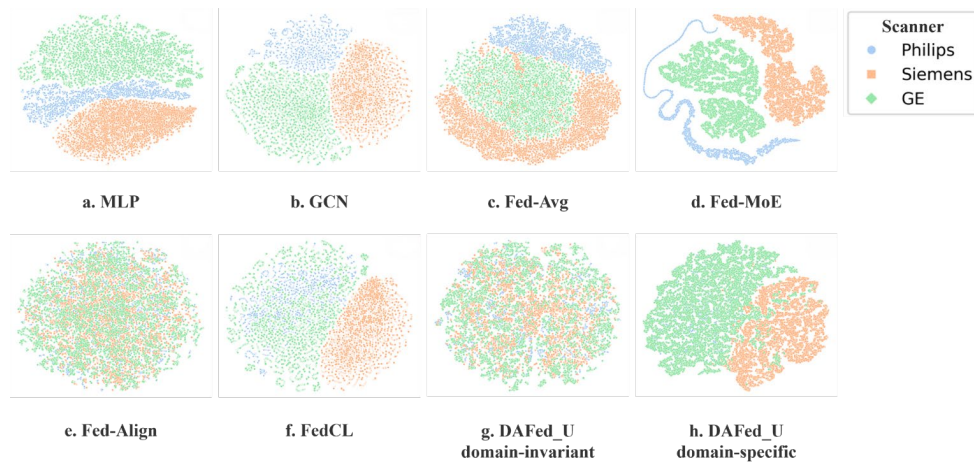

**Fig. 7.** Comparison of latent space visualization by different models on ADNI-3.

Among the four federated learning-based methods, FedAvg achieved an accuracy of 0.686, FedMoE reached 0.649, FedAlign obtained 0.652, and FedCL achieved 0.676. Notably, FedAvg and FedMoE failed to learn robust common domain-invariant features, as illustrated in **Fig. 7c** and **7d**. While FedCL captured some shared features between Philips and Siemens scanners, it could not integrate GE's site-specific features, as shown **in Fig. 7f**. In contrast, FedAlign successfully learned better common domain-invariant features, as demonstrated in **Fig. 7e**. Both non-federated learning methods performed best on the GE dataset, which led us to select GE as the source domain for DAFed. Our approach achieved the highest performance across all three scanners. When utilizing all label information, DAFed_L reached an average classification accuracy of 0.718. In contrast, DAFed_U, which does not rely on label information, achieved the second-highest accuracy of 0.694. As illustrated in **Fig. 7g** and **7h**, DAFed learns robust common domain-invariant features and preserves domain-specific features unique to each scanner.

Through the experimental results on both ABIDE and ADNI-3 datasets, we find that DAfed not only effectively learns robust common domain-invariant features while preserving domain-specific features for each dataset, but also successfully achieves domain adaptation. The model trained on labeled datasets can effectively transfer to the unlabeled data for classification, achieving better performance compared to both federated and non-federated supervised learning methods.

**Ablation Study**

We conduct ablation studies to investigate the role of four key modules in the proposed method. 1) Spatial-Temporal Feature Generator (STFG) module. 2) Representation Disentanglement (RD) module. 3) Domain Adversarial Training (DAT) module and 4) Contrastive Learning (CL) module. As shown in **Table 6**, we validated the contribution of each component in the DAFed method. The results demonstrate that as modules are progressively removed from the DAFed method, the accuracy on the ABIDE and ADNI-3 datasets steadily declines. This indicates that each of the four components plays a crucial role in enhancing classification accuracy, underscoring the significance of every module within the DAFed architecture.

Table 6. Ablation study on model architectures for ABIDE and ADNI-3 Datasets

| Methods | Component | | | | ABIDE | ADNI-3 |
|---|---|---|---|---|---|---|
| | STFG | RD | DAT | CL | Accuracy | Accuracy |
| DAFed | √ | √ | √ | √ | **0.764** | **0.718** |
| w/o C | √ | √ | √ | × | 0.725 | 0.697 |
| w/o CD | √ | √ | × | × | 0.691 | 0.689 |
| w/o CDR | √ | × | × | × | 0.689 | 0.687 |
| w/o CDRS | × | × | × | × | 0.669 | 0.686 |

*w/o C (No CL), w/o CD (No CL and DAT), w/o CDR (No CL, DAT and RD), w/o CDRS (No CL, DAT, RD and STFG)

**Interpretability Assessment**

To highlight the performance of our improved Score-CAM, we adopted evaluation methods that are commonly used by interpretability techniques while extended to graph convolutional networks, such as Grad-CAM [37] and Grad-CAM++ [38], for comparison. We first evaluated the faithfulness of the interpretations produced by Score-CAM for the class recognition task [26]. The Average Drop is defined as:

$$\text{Average Drop} = \sum_{i=1}^{N} \frac{\max(0, Y_i^c - O_i^c)}{Y_i^c} \times 100 \quad (13)$$

The Increase in Confidence (Average Increase) is expressed as:

$$\text{Average Increase} = \sum_{i=1}^{N} \frac{\text{sign}(Y_i^c < O_i^c)}{N} \times 100 \quad (14)$$

where $Y_i^c$ represents the predicted score for class $c$ on sample $i$; $O_i^c$ represents the predicted score for class $c$ when the explanation map region is used as input; sign(·) is an indicator function that returns 1 if the input condition is true.

To demonstrate the capability of our interpretability module, we conducted experiments on the DAFed method using the ABIDE and ADNI-3 datasets. We reported the evaluation results for the 4 graph convolutional layers in our model, as the model leverages outputs from multiple graph convolutional layers in the decision-making process. As shown in **Table 7**, the improved Score-CAM outperforms other explainability methods when extended to GCN in most cases. This suggests that the improved Score-CAM effectively identifies the most distinguishable brain regions in the samples when compared to previous methods, thereby providing a clearer insight into the model's decision-making process.

Table 7. Evaluation results of the improved Score-CAM module on ABIDE and ADNI-3 Datasets

| Dataset | DAFed | Layer | GradCAM | GradCAM++ | Improved ScoreCAM |
|---|---|---|---|---|---|
| ABIDE | Average Drop (%) | 1 | **8.87** | 32.32 | 13.39 |
| | | 2 | 24.13 | 45.24 | **10.82** |
| | | 3 | 44.65 | 42.69 | **12.19** |
| | | 4 | 37.34 | 27.09 | **14.48** |
| | Average Increase (%) | 1 | **50.41** | 25.38 | 35.17 |
| | | 2 | 31.72 | 21.22 | **35.63** |
| | | 3 | 13.86 | 26.19 | **34.5** |
| | | 4 | 21.64 | 30.29 | **33.45** |
| ADNI-3 | Average Drop (%) | 1 | 14.04 | 27.98 | **13.74** |
| | | 2 | 15.17 | 16.38 | **14.21** |
| | | 3 | 13.71 | 27.59 | **13.38** |
| | | 4 | 22.97 | 31.02 | **12.96** |
| | Average Increase (%) | 1 | 43.01 | 43.31 | **45.65** |
| | | 2 | 42.79 | **66.83** | 45.66 |
| | | 3 | 33.99 | 43.88 | **46.35** |
| | | 4 | 40.32 | 36.85 | **45.08** |

## Analysis and Discussion

### Discriminative Brain Region

In this analysis encompassing multi-site data from two cohorts, we have developed a brain functional connectivity analysis based on federated learning model and demonstrated its robustness and reproducibility in the diagnosis of ASD and the early identification of the MCI stage in AD. Our comparative analysis reveals that the specific FC features are strong predictors of ASD and conversion from NC to MCI. Overall, our findings provide compelling evidence of the FC's role in the early identification and prediction of ASD and MCI.

The 111 ROIs in the HO atlas are categorized into seven functional networks [39]: the sensorimotor network (SMN), visual network (VIS), execution and attention network (EAN), default mode network (DMN), subcortical nuclei regions (SBN), limbic system (Limbic), and auditory network (AUD). Similarly, the 116 ROIs in the Automated Anatomical Labeling (AAL) atlas are grouped into six distinct networks [40]: the sensorimotor network (SMN), visual network (VIS), execution and attention network (EAN), default mode network (DMN), subcortical nuclei regions (SBN), and cerebellum (Cer).

**Fig. 8** and **Fig. 9** display the top 10 ROIs with the highest importance scores on ABIDE and ADNI-3, respectively. Using the BrainNet Viewer with the "nearest voxel" mapping algorithm on a standard MNI-space brain surface [40], we plotted these significant ROIs. The scores are calculated by averaging the output values of the improved ScoreCAM across all subjects.

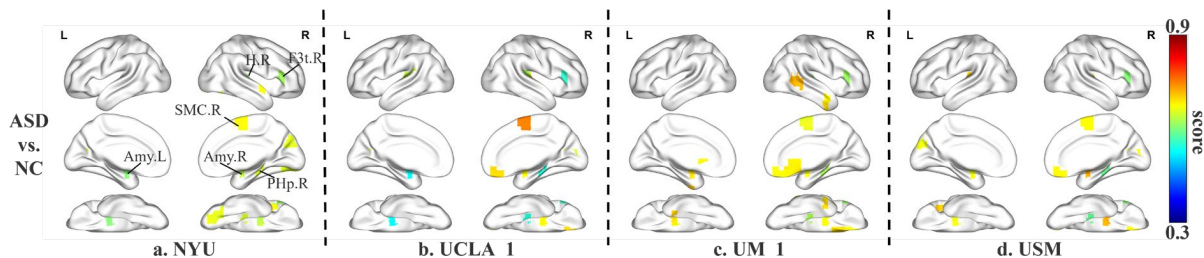

**Fig. 8.** Visualization of the most significant ROIs for ASD using the DAFed method. a – d show the scoring results from the lateral, medial, and ventral views of the brain surface for four different sites.

Our method consistently identified ROIs for ASD versus NC classification on ABIDE across the four sites. As shown in **Fig. 8**, we applied neuroanatomical risk mapping to identify brain regions associated with ASD. The shared ROIs, annotated in **Fig. 8a**, include the Right Juxtapositional Lobule Cortex, part of the supplementary motor area, which has been associated with ASD [41]. The Right Parahippocampal Gyrus, located in the medial temporal lobe, plays a critical role in memory, spatial navigation, and emotion processing, and has also been linked to ASD [6,42]. Additionally, the Right Accumbens, known for its distinct neural activation patterns during reward paradigms in ASD, exhibits notable hypoactivation in the striatum [43]. We also identified the Right Amygdala and Left Amygdala, with studies showing that the Right Amygdala has a significantly larger volume in individuals with ASD [44], and the Left Amygdala shows significantly reduced responses to both threat and safe cues in individuals with ASD [45]. Furthermore, we observed that the Right Planum Polare and Heschl's Gyrus are associated with ASD. Research indicates that these ROIs may correlate with the increased intrapair distance in the SN-OT subnetwork component strength, which in turn correlates positively with subject-specific differences in symptom severity [46]. Lastly, we identified the Right Inferior Frontal Gyrus, which plays a vital role in language processing, comprehension, and sophisticated inhibitory control, and is considered a potential target for therapeutic brain stimulation in individuals with ASD [42,47,48].

Notably, our method also identified site-specific ROIs. For the NYU site, as shown in **Fig. 8a**, the site-specific ROIs identified include the Left Accumbens and the Left Supracalcarine Cortex, both of which have been previously associated with ASD in the literature [42,49]. For the UCLA site, the Right Frontal Medial Cortex, a novel ROI identified in our study, has not been previously reported in the literature to ASD. As a key region of the DMN, the Right Frontal Medial Cortex plays a central role in self-related emotional evaluation, introspective thinking, and social cognition [42]. This ROI was also identified in the UM and USM sites. In **Fig. 8c**, we identified the Right Middle Temporal Gyrus, a key region of the "social brain" network, which has been reported to be associated with ASD [50]. For the USM site, we discovered the Left Frontal Operculum Cortex, which has not yet been reported to ASD. This region is part of the language network and plays a critical role in language processing. It is also involved in cognitive control and attention regulation, making it a subject worthy of further investigation.

Similarly, in **Fig. 9**, we applied neuroanatomical risk mapping to identify significant ROIs for MCI versus NC classification across Philips, Siemens, and GE scanners. The common ROIs discovered across the three scanners were annotated in **Fig. 9a**, including the Superior Frontal Gyrus, Left Heschl's Gyrus, Left Angular Gyrus, and Lobules I and II of the Vermis. Metabolic alterations in the Superior Frontal Gyrus have been validated as being associated with cognitive decline. Similarly, metabolic changes in the Superior Frontal Gyrus have been observed in individuals with MCI who later develop AD [51]. Hyperactivation in regions such as Heschl's Gyrus and the Insula has also been found in MCI patients [52]. Additionally, the Left Angular Gyrus and Lobules I and II of the Vermis have been implicated from the MCI stage, with subsequent involvement of the hemispheric part of the posterior lobes and Crus I observed exclusively in AD patients [53, 54]. These findings are consistent with our study, indicating that functional ROI features may play a role in the early identification of cognitive decline and in predicting progression to AD.

Our method also revealed some interesting results across different MRI devices. On the Philips scanner, as shown in **Fig. 9a**, we found that the Left Inferior Frontal Gyrus, Right Insula, Right Anterior Cingulate and Paracingulate Gyri are associated with MCI. Previous studies have reported the Left Inferior Frontal Gyrus and Right Insula to MCI [55, 56]. The Right Anterior Cingulate and Paracingulate Gyri, part of the DMN, plays a critical role in higher-order cognitive control and represents a novel finding in our study. Similarly, in **Fig. 9b**, the Left Anterior Cingulate and Paracingulate Gyri are also a new finding, along with the Right Insula and Right Superior Temporal Gyrus, which have been validated as being associated with MCI [56, 57]. Finally, on the GE scanner, we identified the Left Gyrus Rectus and Lobules IV and V of the Vermis. The Left Gyrus Rectus, part of the DMN, is involved in higher-order functions such as cognitive control and reward processing and has been reported to be associated with MCI [58]. Although there is no direct evidence linking Lobules IV and V of the Vermis to MCI, recent studies suggest they are widely involved in motor functions, neurodevelopment, and neuropsychiatric disorders [59].

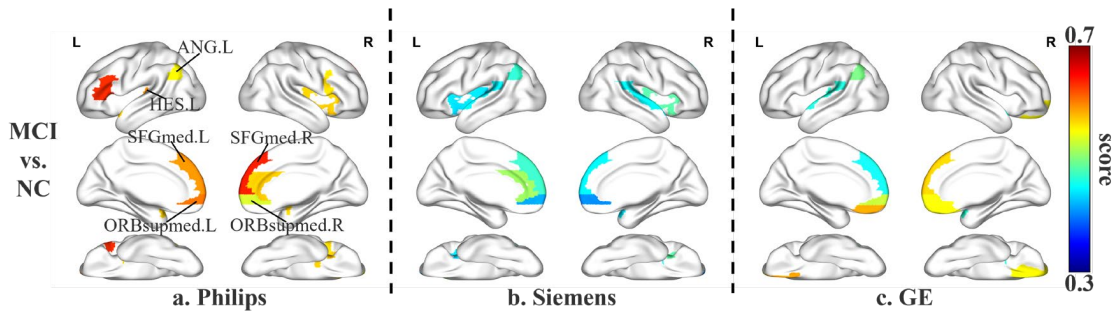

**Fig .9**. Visualization of the most significant brain biomarkers for MCI using the DAFed method. a – c display the scoring results from the lateral, medial, and ventral views of the brain surface for three scanners.

Based on the improved ScoreCAM scores across all subjects, a comparison of the score distribution between the patient and NC groups reveals that the identified ROIs exhibit strong discriminatory power. As shown in **Fig. 10**, the violin plots are arranged in descending order of the mean absolute score values for the corresponding ROIs. Blue (ASD) and orange (NC) represent the two groups' distributions across various ROIs.

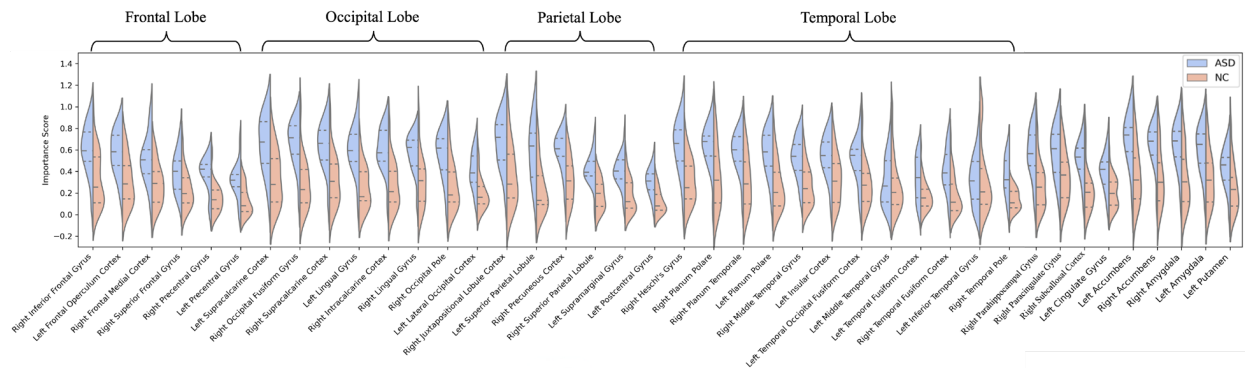

**Fig. 10**. A violin plot illustrating the distribution of ROI scores for the ASD and NC groups.

In the Frontal Lobe, particularly within the Inferior Frontal Gyrus and Frontal Operculum Cortex, importance scores are concentrated in the higher range. This highlights the close association between abnormalities in the prefrontal cortex and impairments in language, emotional and social cognition, as well as motor coordination in individuals with ASD [60].

Abnormalities in the Occipital Lobe, such as the Left Supracalcarine Cortex, Right Occipital Fusiform Gyrus, and Lingual Gyrus, *et al*., may reflect disruptions in visual processing in individuals with ASD [61].

Significant differences have been observed in the Parietal Lobe, particularly in regions involved in sensory integration, spatial cognition, self-awareness, and attention. Notable areas include the Right Juxtapositional Lobule Cortex, Superior Parietal Lobule and Precuneus Cortex.

Several ROIs in the Temporal Lobe exhibit varying distribution differences in individuals with ASD, particularly in regions such as the Middle Temporal Gyrus, Temporal Pole, and Planum Temporale, associated with language, social interaction, and emotional processing. These differences may underlie the core symptoms of ASD, including social deficits, language development delays, and sensory abnormalities [62].

ASD patients also exhibit significant distribution differences across multiple brain regions involved in emotion, cognition, and social function. Notable ROIs, such as the Amygdala, Cingulate Gyrus, and Accumbens, further highlight their importance in the pathophysiology of ASD.

**Fig. 11** also illustrates the distribution of ROI scores for the MCI and NC groups. Specifically, the Inferior Frontal and Middle Frontal Gyrus within the Frontal lobe are involved in executive function, decision-making, and language abilities. These regions may exhibit compensatory enhancements or abnormal activity in individuals with MCI. The Paracentral Lobule, linked to sensorimotor functions, shows differences that may be related to mild impairments in spatial perception or motor coordination in individuals with MCI [63]. The Supplementary Motor Area, associated with motor planning and cognitive control, may show abnormal activity in MCI patients, reflecting early signs of declining cognitive control [64].

In the Occipital Lobe, ROIs such as the Lingual Gyrus, the Inferior Occipital Gyrus, the Calcarine and the Fusiform are significantly associated with MCI. These areas are involved in visual processing, memory, and cognitive integration. MCI may lead to a decline in memory, information processing, and attention by affecting the visual cognition-related regions [65].

In the Parietal Lobe, ROIs such as the Angular Gyrus, the Parietal Superior and the Parietal Inferior show significant differences between the MCI group and the NC group. These areas are associated with language, attention, and spatial cognition. Functional changes in the parietal regions reflect impairments in higher cognitive functions in MCI patients, such as attention, memory retrieval and language integration. These impairments may be caused by brain atrophy or a decline in connectivity due to neurodegenerative diseases [66].

Temporal Lobe atrophy is a hallmark of MCI. Heschl's gyrus, which plays a critical role in auditory processing, often exhibits functional abnormalities in MCI patients, potentially contributing to declines in language processing abilities. The Superior Temporal Gyrus is essential for language comprehension, with significant functional reductions observed in the Left Superior Temporal Gyrus. Impairment in semantic memory may be a key factor underlying language comprehension deficits in MCI patients. In addition, the Middle and Superior Temporal Gyri serve as crucial nodes within the DMN. Dysfunction in these areas may disrupt network connectivity, further affecting memory and cognitive functions [67]. Functional abnormalities in the Inferior Temporal Gyrus are frequently associated with deficits in memory and visual-semantic processing [68], which could explain the decline in visual memory and cognitive abilities commonly seen in MCI patients.

Moreover, abnormalities in other ROIs, such as the Vermis, may be associated with emotional and attentional regulation deficits in MCI patients. Weakened functional connectivity between the Cerebellum, Frontal Lobe and limbic system may contribute to cognitive decline. Reduced activity in the Insula could lead to impairments in emotional regulation and social cognition, aligning with the emotional instability and behavioral changes commonly observed in MCI. The Cingulum Ant, a key region for attentional control and emotional responses, is closely linked to the DMN. Dysfunction of DMN nodes may serve as an early indicator of MCI [69]. Changes in the Parahippocampal, Cingulum and Amygdala have been identified as biomarkers of AD, alterations in these regions may help predict whether MCI will advance to AD [70,71].

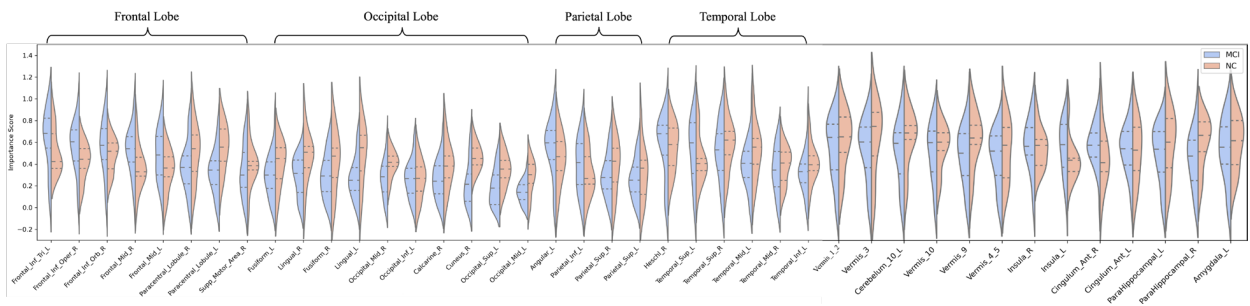

**Fig. 11**. A violin plot illustrating the distribution of ROI scores for the MCI and NC groups.

**Discriminative Functional Connectivity**

The Pearson correlation was calculated with a two-tailed p-value, and edges with p-values greater than 0.05 were excluded. We retained the top 10 significant ROIs with the largest absolute correlation values among all ROIs to highlight the considerable differences in connection strength between patients and healthy controls.

**Fig. 12** visualizes the significant FCs between ASD and NC across the four sites of ABIDE. Specifically, **Fig. 12e** shows the FCs shared across these four sites. We observed that ASD exhibited increased connectivity between networks such as the SBN, EAN, and AUD, often interpreted as pathological crosstalk between networks. This could lead to inefficiencies in information processing and functional conflicts [72]. In addition, at the UCLA site, we found that ASD exhibited reduced connectivity within the VIS network compared to NC and increased local connectivity between the DMN and VIS. At the UM site, ASD showed increased connectivity between the DMN and AUD networks and decreased connectivity within the DMN network compared to NC. At the USM site, ASD exhibited enhanced connectivity between DMN and EAN, as well as between the DMN and AUD, compared to NC. These findings align with those reported by Morgan et al. [73], who characterized ASD by reduced intra-network connectivity and increased inter-network connectivity. Additionally, Washington et al. [74] found that children with ASD exhibit reduced connectivity between DMN nodes and increased local connectivity between DMN nodes and visual networks. Our findings further support these observations.

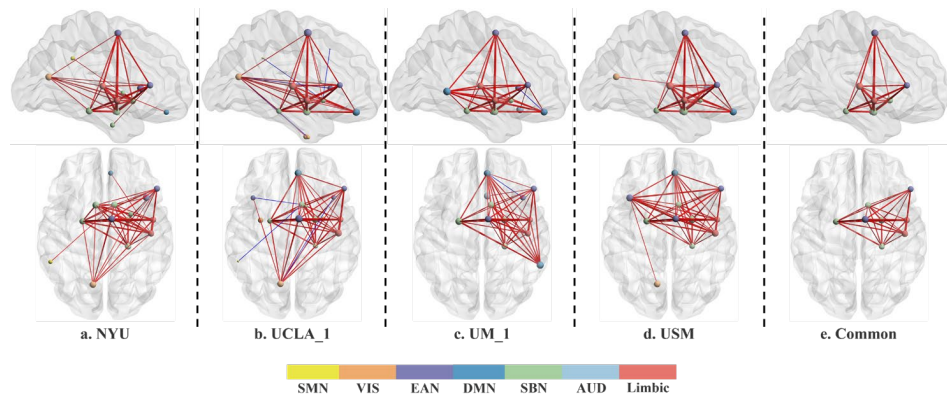

**Fig. 12.** The significant functional connectivity in the ABIDE dataset.

**Fig. 13** visualizes significant FCs between the important ROIs for MCI and NC. **Fig. 13d** highlights the common FCs across the three scanners. Compared to NC, MCI exhibited reduced FC within the DMN, such as the connection between the Superior Frontal Gyrus and the Angular Gyrus. This observation highlights a potential pathological feature of MCI, where reduced FC within the DMN might contribute to deficits in memory or other cognitive functions. Previous studies have indicated that DMN regions overlap with areas of amyloid plaque deposition [75], providing valuable insight into the neural mechanisms underlying MCI. In addition, across different devices, we identified several abnormal FCs. As shown in **Fig. 13a** and **13b**, MCI exhibited reduced local connectivity within the DMN, such as the connection between the Angular Gyrus and the Parahippocampal Gyrus. These regions are key areas involved in amyloid plaque deposition and the disruption of DMN connectivity [76]. Similarly, in **Fig. 13c**, we observed a reduction in some local connectivity within the DMN. The discovery of these scanner-specific features and FCs is partly attributed to data heterogeneity caused by differences between devices, and partly due to variations in sample size across different scanners. These findings offer valuable insights from multiple perspectives and enrich our understanding of MCI from the functional brain region viewpoint.

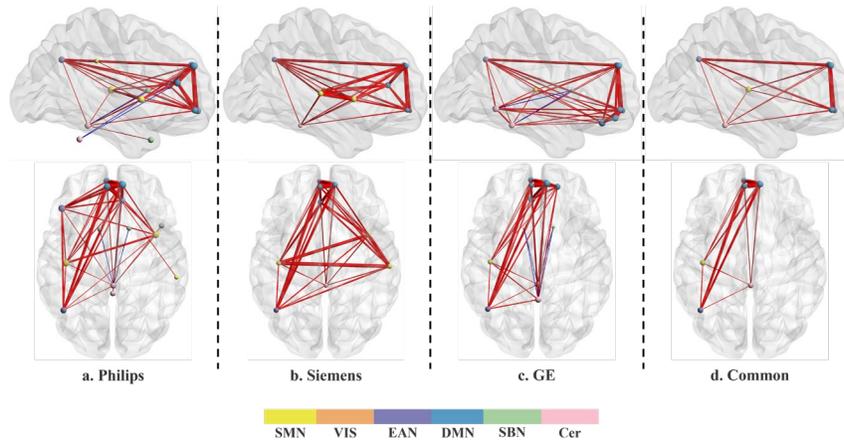

**Fig.13.** The significant functional connectivity in the ADNI dataset.

In conclusion, we proposed a Domain Adversarial Federated Learning framework to collaboratively trains model across multiple sources without data sharing. Our method employs a feature disentanglement module to decompose latent features generated by the feature generator into domain-invariant and domain-specific components. Through adversarial training in a federated learning framework, the proposed method ensures that multiple datasets learn shared domain-invariant features and enables to transfer the model parameters learned from a labeled dataset to an unlabeled one. In addition, contrastive learning is applied to self-supervised learning of domain-invariant features in the unlabeled dataset.

The proposed method successfully classified ASD and MCI across the ABIDE and ADNI datasets, demonstrating the generalizability of the federated learning framework. Furthermore, we utilized an improved Score-CAM module that highlighted several ROIs and FCs associated with two disease models, i.e., the neurodevelopmental disorder ASD and the neurodegenerative disease AD, offering a functional perspective on these conditions.

Overall, these findings suggest that the identified fMRI features provide valuable insights into distinctive characteristics of brain subregions in individuals with ASD and MCI, thereby enhancing the understanding and characterization of the underlying pathophysiological mechanisms associated with ASD and cognitive impairment in AD.

**Acknowledgements**

This research was supported in part by the National Center for Advancing Translational Research of NIH under Grant UM1TR004771; and in part by NIH under Grant P20GM109036, U19AG055373, R01AG061917, and R01AG068232.